\documentclass[conference]{IEEEtran}
\IEEEoverridecommandlockouts
\usepackage{cite}
\usepackage{url}
\usepackage{amsmath,amssymb,amsfonts}
\usepackage{algorithmic}
\usepackage{graphicx}
\usepackage{textcomp}
\usepackage{xcolor}
\usepackage{amssymb}
\usepackage{subfigure}
\def\BibTeX{{\rm B\kern-.05em{\sc i\kern-.025em b}\kern-.08em
    T\kern-.1667em\lower.7ex\hbox{E}\kern-.125emX}}

\begin{document}

\title{A Robust Deep Unfolded Network for Sparse Signal Recovery from Noisy Binary Measurements}\author{\IEEEauthorblockN{Yuqing Yang$^\ddagger$, Peng Xiao$^{\ddagger\star}$, Bin Liao$^{\star}$, Nikos Deligiannis$^{\ddagger\dagger}$}
\IEEEauthorblockA{$^\ddagger$\textit{Department of Electronics and Informatics, Vrije Universiteit Brussel, Pleinlaan 2, B-1050 Brussels, Belgium} \\
$^\star$\textit{Department of Electronics and Information Engineering, Shenzhen University, China}\\
$^\dagger$\textit{imec, Kapeldreef 75, B-3001 Leuven, Belgium}\\
{Email: yuqing.yang@vub.be, pxiao@etrovub.be, binliao@szu.edu.cn, ndeligia@etrovub.be}
}}\vspace{-1em}



\maketitle

\begin{abstract}
We propose a novel deep neural network, coined DeepFPC-{$\ell_2$}, for solving the 1-bit compressed sensing problem. The network is designed by unfolding the iterations of the fixed-point continuation (FPC) algorithm with one-sided $\ell_2$-norm (FPC-$\ell_2$). The DeepFPC-$\ell_2$ method shows higher signal reconstruction accuracy and convergence speed than the traditional FPC-$\ell_2$ algorithm. Furthermore, we compare its robustness to noise with the previously proposed DeepFPC network---which stemmed from unfolding the FPC-$\ell_1$ algorithm---for different signal to noise ratio (SNR) and sign-flipped ratio (flip ratio) scenarios. We show that the proposed network has better noise immunity than the previous DeepFPC method. This result indicates that the robustness of a deep-unfolded neural network is related with that of the algorithm it stems from.

\end{abstract}


\section{Introduction}
Compressive sensing (CS) is a low-rate signal acquisition method for sparse signals\cite{donoho2006compressed}. Provided that a signal is sparse, or has a sparse representation in a set of bases or learned dictionary, CS can recover it with a sampling rate notably lower than what the Shannon-Nyquist theory dictates. CS has found many applications in image processing\cite{romberg2008imaging}, wireless communications\cite{bajwa2006compressive} and direction-of-arrival (DOA) estimation\cite{malioutov2005sparse}, to name a few. In practice, the measurements of CS must be quantized to discrete values; 1-bit CS\cite{boufounos20081}, in particular,
considers 1-bit quantized measurements. Compared with traditional CS, 1-bit CS can further reduce the sampling and  system complexity, i.e., by realizing the sampling with a simple comparator. Yet, a challenge for the 1-bit CS problem is how to solve the related optimization problem with high accuracy. 

A variety of algorithms have been proposed to solve the 1-bit CS problem. The fixed point continuation (FPC) method in\cite{hale2007fixed} can be deployed by computing the one-sided $\ell_1$- or $\ell_2$-norm penalty on the unit sphere $\|\boldsymbol{x}\|_{2}=1$ and applying a re-normalization step, resulting in the FPC-based algorithms FPC-$\ell_1$\cite{xiao20191} and FPC-$\ell_2$\cite{boufounos20081}, respectively. Alternatively, the matching sign pursuit (MSP)\cite{boufounos2009greedy}, the binary iterative hard thresholding (BIHT)\cite{jacques2013robust} and the History algorithm\cite{sun2016history} have been proposed, achieving high reconstruction performance. FPC-based algorithms do not require knowing the level of sparsity in advance, knowledge that might not be available in a practical setting. However, the performance of FPC-based algorithms is relatively poor compared to other algorithms that need the sparsity level as an input. Thus, a more accurate reconstruction method that does not require knowledge on the signal sparsity level is desired.


Deep neural networks (DNNs) have demonstrated state-of-the-art performance in many inference problems \cite{mansanet2016local,snoek2015scalable}, including  signal recovery tasks~\cite{nguyen2017deep}.
Nevertheless, DNNs are considered as black-box models and do not incorporate prior knowledge about the signal structure. Deep unfolding\cite{gregor2010learning} is expected to bridge the gap between analytical methods and deep learning-based methods by designing DNNs as unrolled iterations of optimization algorithm. Deep unfolding networks are interpretable (as opposed to traditional DNNs) and have proven to be superior to traditional optimization-based methods and DNN models (because they integrate prior knowledge about the signal structure). Examples of such networks include LISTA\cite{gregor2010learning}, 
{$\ell_1$}-{$\ell_1$}-RNN\cite{le2019designing}, and LeSITA\cite{tsiligianni2019deep}.
In a recent study we presented a DeepFPC network\cite{xiao2019deepfpc} designed by unfolding the iterations of the FPC-{$\ell_1$} algorithm. DeepFPC achieved good performance in solving the 1-bit CS problem without requiring knowing the signal sparsity level. It was, however, shown that the reconstruction performance of DeepFPC~\cite{xiao2019deepfpc} degrades when the measurements are contaminated by high noise. This calls for a robust deep unfolded network showing improved performance in the presence of noise. 


The contribution of this paper is as follows: ($i$) we design a novel DeepFPC-{$\ell_2$} network by corresponding each layer of an unfolded deep network to an iteration of the FPC-\textit{$\ell_2$} algorithm; ($ii$) we propose extended matrices to reduce the time complexity and improve the speed and efficiency of the designed network; ($iii$) we compare the robustness of the previous DeepFPC~\cite{xiao2019deepfpc} and the proposed DeepFPC-{$\ell_2$} for different signal to noise ratios (SNR) and sign-flipped to signal ratios (flip ratio), and shows that the latter is more robust to noise. This indicates that the robustness of a unfolded network is related to the that of the algorithm it stems from.

In what follows, the next section provides a brief background on the FPC-based method and compares the robustness to noise of the one-sided $\ell_1$-norm and the one-sided $\ell_2$-norm formulations. 
Section \uppercase\expandafter{\romannumeral3} introduces the proposed DeepFPC-{$\ell_2$} model and Section \uppercase\expandafter{\romannumeral4} reports simulation results. 
Finally, Section \uppercase\expandafter{\romannumeral5} concludes the paper. 

\textit{Notation}: Throughout the paper, $\boldsymbol{y} = [y_1,y_2,...,y_M]^T \in {\{}{-}{1}{,}{1}{\}}^{M}$ is the vector of measurements, $\boldsymbol{{\Phi}} \in {\mathbb{R}}^{M \times N}$ is the sensing matrix, and $\boldsymbol{x} \in {\mathbb{R}}^{N}$ is the sparse signal with sparsity $K \ll N$. Moreover, $\tilde{\boldsymbol{X}}=[\textbf {x}_1, \boldsymbol {x}_2,\dots,\boldsymbol {x}_L]\in \mathbb{R}^{N\times L}$ combines $L$ sparse signal vectors into a signal matrix, $\tilde{\boldsymbol{Y}}=[\boldsymbol {y}_1, \boldsymbol {y}_2,\dots,\boldsymbol {y}_L]\in \mathbb{R}^{M\times L}$ is the measurement matrix combining $L$ measurement vectors and $\boldsymbol{Y}_l=\text{diag}(\boldsymbol{y}_l)$ is a square matrix with $\boldsymbol{y}_l$ being the diagonal entries.  Furthermore, $\odot$ and $\otimes$ respectively denote the matrix-multiply product and the Hadamard product, $\left(\cdot\right)^T$ is the transpose operation, and $\|\cdot\|_p$ represents the $p$-norm and $|\cdot|$ is the element-wise absolute operator.

\begin{figure}[t]
\centering
\includegraphics[scale = 0.4]{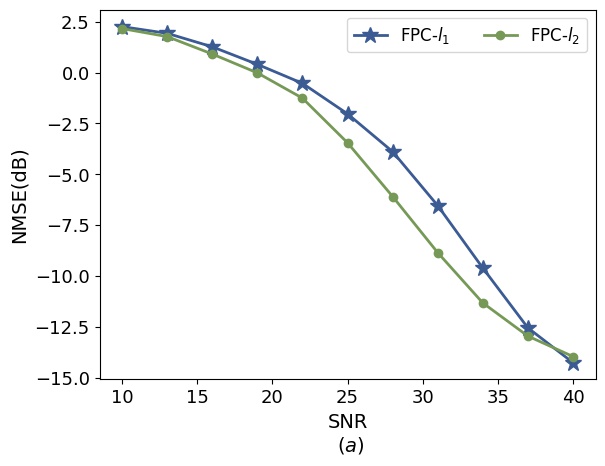}
\hspace{0.25cm}
\includegraphics[scale = 0.4]{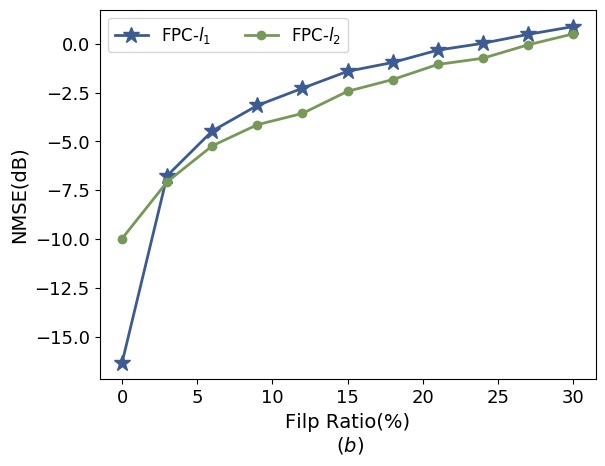}
\caption{Reconstruction NMSE for FPC-{$\ell_1$} and FPC-{$\ell_2$}; the sign flips of measurements are caused by (a)  measurement noise; (b) transmission errors.}
\label{fig: algorithm}
\end{figure}

\begin{figure*}[htbp]
\centerline{\includegraphics[scale = 0.5]{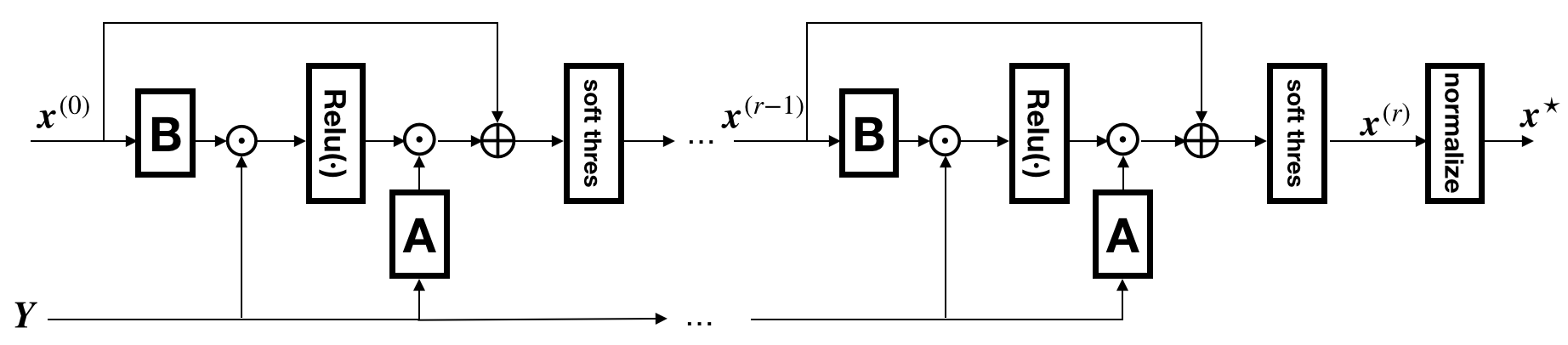}}
\caption{The block diagram of proposed DeepFPC-{$\ell_2$} network.}
\label{fig.1}
\end{figure*}

\section{Background on FPC-based Methods} 
In this section, two FPC-based algorithms, i.e., FPC-{$\ell_1$} and FPC-{$\ell_2$}, are introduced.
Moreover, we review their robustness to noise, which forms our motivation to design the DeepFPC-{$\ell_2$} model.

\subsection{The FPC-{$\ell_1$} and  FPC-{$\ell_2$} Algorithms}
In 1-bit CS, the measurements are acquired as:
\begin{equation}\label{measurement}
    \boldsymbol{y} = \text{sign}(\boldsymbol{{\Phi}}\boldsymbol{x})
\end{equation}
where the $\textnormal{sign}(\cdot)$ function is applied element-wise to the vector $\boldsymbol{{\Phi}}\boldsymbol{x}$. The FPC-based algorithm for 1-bit CS recovery solves the following problem:
\begin{equation}\label{problem}
    \hat{\boldsymbol{x}} = \mathop{\mathrm{argmin}}\limits_{\boldsymbol{x}}{\|{\boldsymbol{x}}\|}_{1}{+}\lambda\mathop{\sum}\limits_{{i}{=}{1}}\limits^{M}{{f}{((}{\boldsymbol{Y}}\boldsymbol{{\Phi}}{\boldsymbol{x}}{)}_{i}{)}\hspace{0.5em}{\mathrm{s}}{.}{\mathrm{t}}{.}\hspace{0.5em}}{\|{\boldsymbol{x}}\|}_{2}{=}{1}
\end{equation}
where $\lambda$ is a tuning parameter ($\lambda > 0$) 
and $f(x)$ can be specified as the one-sided $\ell_1$-norm:
\begin{equation}
f_{\ell_1}(x)=
\begin{cases}
    0, & x\geq 0\\
    \lvert x\rvert, & x< 0
\end{cases}
\end{equation}
or the one-sided {$\ell_2$}-norm penalty: 
\begin{equation}
f_{\ell_2}(x)=
\begin{cases}
    0, & x\geq 0\\
    \frac{x^2}{2}, & x< 0
\end{cases}
\end{equation}

The convexity  and smoothness of this function allows the use of FPC method to perform the minimization. The FPC method solves Problem (\ref{problem}) by updating $\boldsymbol{x}$ iteratively as:
\begin{subequations}\label{5a}
\begin{align}
\label{4-1}
& {\boldsymbol{u}}=S_{\nu}\left(\boldsymbol{x}^{(r)}-\tau\mathbf{g}(\boldsymbol{x}^{(r)})\right),
\\
\label{4-2}
& \boldsymbol{x}^{(r+1)}={{\boldsymbol{u}}}/{\| {\boldsymbol{u}}\|_2},
\end{align}
\end{subequations}
where $\tau > 0$, $\nu = \tau/\lambda$ and ${{S}}_{{\nu}}(\cdot) \triangleq \textnormal{sign}(\cdot)\odot\mathrm{max}{\lbrace|\cdot|-\nu, 0{\rbrace}}$ is the soft-thresholding operator. Furthermore, $g(\boldsymbol{x})$ is the gradient of the consistent term. When the one-sided {$\ell_1$}-norm is considered, the gradient is given by
\begin{equation}
        g_{\ell_1}(\boldsymbol{{x}}) = {{\boldsymbol{{\Phi}}}^{\mathrm{T}}}{(\mathrm{sign}({\boldsymbol{{\Phi}}}{\boldsymbol{x})}-\boldsymbol{y})},
\end{equation}
whereas, when the one-sided {$\ell_2$}-norm is considered, the gradient becomes:
\begin{equation}\label{gradient}
    g_{\ell_2}(\boldsymbol{{x}}) = {{\boldsymbol{{\Phi}}}^{\mathrm{T}}}{\boldsymbol{Y}}{\bar{{f_{\ell_2}^\prime}}}({\boldsymbol{Y}}{\boldsymbol{{\Phi}}}{\boldsymbol{x}}),
\end{equation}
where $\boldsymbol{Y} = \mathrm{diag}(\boldsymbol{y})$ and
\begin{equation}\label{f_h}
\bar{f_{\ell_2}^\prime}(x) =
\begin{cases}
    0, & x> 0\\
    x, & x\leq 0
\end{cases}
\end{equation}

Given proper values for $\tau$, $\lambda$ and initialization $\boldsymbol{{x}}^{({0})}$, the sparse signal can be reconstructed by these two FPC-based algorithms.

\subsection{The FPC-{$\ell_1$} and FPC-{$\ell_2$} Under Noisy Measurements}

Due to noise or transmission errors,  the binary measurements could suffer from sign flips, which will degrade the reconstruction performance of 1-bit CS. To address this issue, a robust recovery method is required. The authors of~\cite{jacques2013robust} showed that the BIHT-$\ell_2$ algorithm outperforms the BIHT-{$\ell_1$} algorithm when the measurements contain sign flips. This means that the one-sided {$\ell_2$} objective has better robustness against noise than the one-sided {$\ell_1$}. We therefore conjecture that the FPC-{$\ell_2$} algorithm has better noise robustness than the FPC-{$\ell_1$} algorithm. To test this hypothesis, we compare their performance for two kinds of sign flips (in the measurement process and in the transmission process) and show the results in Fig.\ref{fig: algorithm}. It can be seen that FPC-{$\ell_2$} has indeed better recovery performance than FPC-{$\ell_1$}. Thus, we propose to design an DeepFPC-{$\ell_2$} network by unfolding the FPC-{$\ell_2$} algorithm to improve the recovery performance under noise. 

\section{The Proposed DeepFPC-{$\ell_2$} Model}

In this section, we present our DeepFPC-{$\ell_2$} architecture (Section \ref{3A}) and describe how to build extended matrices for parallel computing (Section \ref{3B}).   

\subsection{The Proposed Deep Unfolded Network Architecture}
\label{3A}
Substituting $g_{\ell_2} (\boldsymbol{x})$ in (\ref{gradient}) into  the iteration (\ref{4-1}), the update becomes:
\begin{equation}\label{ori}
    \boldsymbol{u}\hspace{0.33em}{=}\hspace{0.33em}{{S}}_{{\nu}}{(}\boldsymbol{{x}}^{({r})} - \boldsymbol{A}\boldsymbol{Y}{\bar{{f_{\ell_2}^\prime}}}(\boldsymbol{Y}\boldsymbol{B}{\boldsymbol{x}}))
\end{equation}
where $\boldsymbol{A} = {\tau}{{\boldsymbol{{\Phi}}}^{\mathrm{T}}}$ and $\boldsymbol{B} = \boldsymbol{{\Phi}}$. Following the principles in \cite{gregor2010learning}, the update can be presented in the form of neural-network layers. An initialization $\boldsymbol{{x}}^{({0})}$ and a diagonal matrix containing the binary measurement vector $\boldsymbol{Y}$ are as inputs to the network. The matrices $\boldsymbol{A}$ and $\boldsymbol{B}$ are trainable linear weights, which may change per layer. Furthermore, ${{S}}_{{\nu}}(\cdot)$ and ${\bar{{f_{\ell_2}^\prime}}}(\cdot)$ play the role of non-linear activation functions in the network. 

\begin{figure}[t]
\centerline{\includegraphics[scale = 0.4]{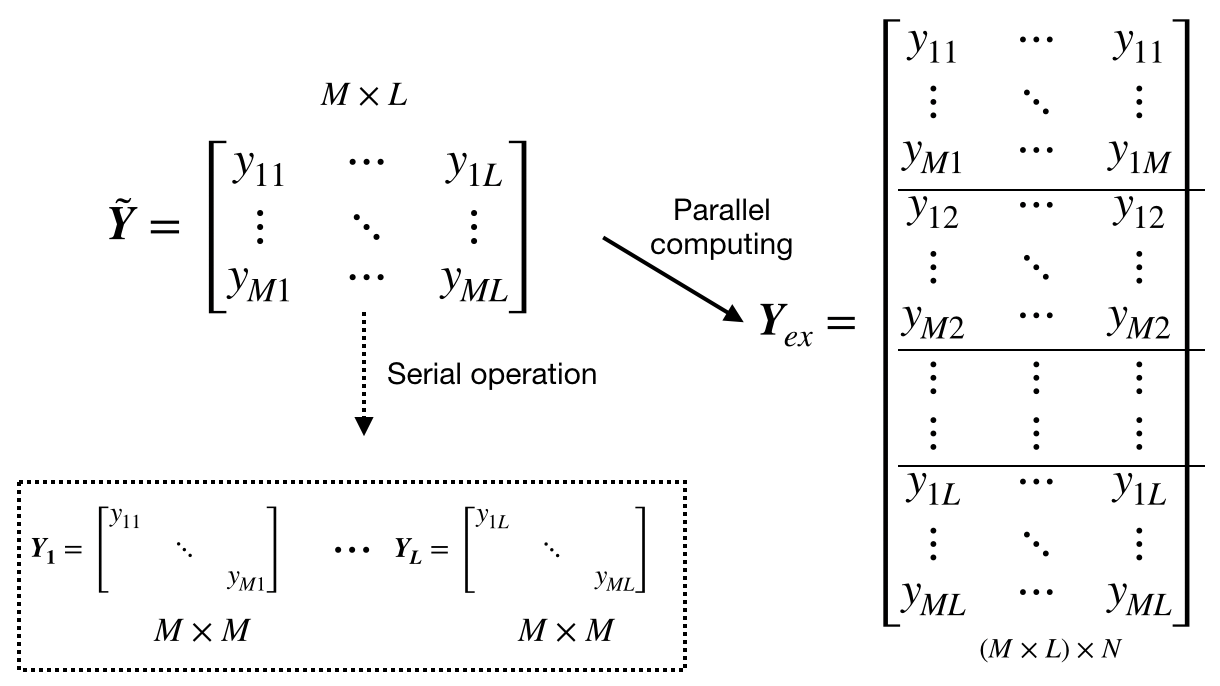}}
\caption{Comparison of the serial and parallel operation.}
\label{fig.4}
\end{figure}

Unlike the traditional FPC-{$\ell_2$} algorithm, the DeepFPC-{$\ell_2$} network does not need to apply a normalization step after every layer, because the network is trained with signal examples and it will not learn to generate zero solutions of finite layers. Hence, normalization is performed after the final layer. Moreover, ${\bar{{f_{\ell_2}^\prime}}}(\cdot)$ in \eqref{f_h} can be expressed as the $\text{relu}(\cdot)$ function\cite{ramachandran2017searching}, which leads \eqref{ori} to:
\begin{equation}\label{8}
   \boldsymbol{x}^{({r+1})}\hspace{0.33em}{=}\hspace{0.33em}{{S}}_{{\nu}}{(}\boldsymbol{{x}}^{({r})} + \boldsymbol{A}\boldsymbol{Y}{\text{relu}}({\boldsymbol{Y}}\bar{\boldsymbol{B}}{\boldsymbol{x}^{(r)}}))
\end{equation}
where $\bar{\boldsymbol{B}} = -\boldsymbol{B} = -\boldsymbol{{\Phi}}$. In the last layer, the final estimate is normalized as ${\boldsymbol{x}^{\star}} = {\boldsymbol{x}^{(R)}}/ {{\left\|{\boldsymbol{x}^{(R)}}\right\|}_{2}}$. 

The block diagram of the proposed DeepFPC-{$\ell_2$} is shown in Fig.\ref{fig.1}. Specifically, the input to the soft-thresholding non-linearity is fed through a shortcut connection from the output of the previous layer. This implies that DeepFPC-{$\ell_2$} can undergo deep training. 
\subsection{Built with Extended Matrices}\label{3B}

As illustrated in \eqref{8} and Fig. \ref{fig.1}, the input $\boldsymbol{Y}$ of the DeepFPC-$\ell_2$ network is the diagonal matrix of one measurement vector $\boldsymbol{y}$. 
For the measurement matrix $\tilde{\boldsymbol{Y}}$, each column, i.e., one measurement vector $\boldsymbol{y}_l$, has to be diagonalized firstly as $\boldsymbol{Y}_l$  (shown in Fig.\ref{fig.4}, serial operation). Consequently, $\boldsymbol{x}_l$ is recovered from each input $\boldsymbol{Y}_l$ one by one, and the operators ($\boldsymbol{A}\boldsymbol{Y}_l$ and $\boldsymbol{Y}_l\bar{\boldsymbol{B}}\boldsymbol{x}$) have to be computed multiple times, which is quite time-consuming. 
Hence, we aim to use parallel computing to reduce time complexity and improve the speed and efficiency of the proposed network.

Our strategy is to expand the matrix to perform parallel computing. Firstly, $\boldsymbol{Y}_{ex}$ is designed by extending the measurement matrix $\tilde{\boldsymbol{Y}}$ (shown in Fig. \ref{fig.4} as \textit{parallel computing}) and then used it as the network input. 
At the same time, Hadamard product operations are used between extended matrices, instead of the matrix-multiply product. 
In this way, we can still get the same result 
with paralleled sum and reshape operations. Specifically, considering that $\boldsymbol{Y}_l$ is a diagonal matrix, we can write:
\begin{equation}
\begin{aligned}
\boldsymbol{Y}_l{\boldsymbol{B}}&=[y_{1l}{\boldsymbol{ b}}^T_1, y_{2l}{\boldsymbol{ b}}^T_2,\dots,y_{Ml}{\boldsymbol{ b}}^T_M]^T\\
&=[\hat{\boldsymbol{y}}^T_{l1}\otimes{\boldsymbol{ b}}^T_1,\hat{\boldsymbol{y}}^T_{l2}\otimes{\boldsymbol{ b}}^T_2,\dots,\hat{\boldsymbol{y}}^T_{lM}\otimes{\boldsymbol{ b}}^T_M]^T\\
&={\hat{\boldsymbol{Y}}_l}\otimes {\boldsymbol{B}},
\end{aligned}
\end{equation}
where $\boldsymbol{b}_m$ is the $m$-th row vector of $\boldsymbol{B}$, $\hat{\boldsymbol{y}}_{ml}$ is an $N$-dimensional row vector with all entries equal to $y_{ml}$, and ${\hat{\boldsymbol{Y}}_l}=\big[{\hat{\boldsymbol{y}}_{1l}}^T,{\hat{\boldsymbol{y}}_{1l}}^T,\dots,{\hat{\boldsymbol{y}}_{1l}}^T\big]^T$. This implies that the matrix-multiply product is replaced by the Hadamard product. We define the following extended matrices as $\boldsymbol{Y}_{ex}=\big[{\hat{\boldsymbol{Y}}_1}^T,{\hat{\boldsymbol{Y}}_2}^T,\dots,{\hat{\boldsymbol{Y}}_N}^T\big]^T\in \mathbb{R}^{(M\times L)\times N}$ and $\boldsymbol{B}_{ex}=\big[{{\boldsymbol{B}}}^T,{{\boldsymbol{B}}}^T,\dots,{{\boldsymbol{B}}}^T\big]^T\in \mathbb{R}^{(M\times L)\times N}$, i.e., we reshape $\tilde{\boldsymbol{Y}}$ into a column vector and copy it $N$ times horizontally to build the extended matrix $\boldsymbol{Y}_{ex}$. Moreover, we copy ${\boldsymbol{B}}$ $L$ times vertically to build the  extended matrix $\boldsymbol{B}_{ex}$. Then, we can compute all $\boldsymbol{Y}_l{\boldsymbol{B}}$ at the same time by a Hadamard product $\boldsymbol{Y}_{ex}\otimes\boldsymbol{B}_{ex}$, as shown in \eqref{6}.


Similarly, $\boldsymbol{X}_{ex} \in \mathbb{R}^{N \times (M\times L)}$ is generated by copying each sample of $\boldsymbol{x}_l$ horizontally $M$ times [see \eqref{7}], and $\boldsymbol{A}_{ex}\in \mathbb{R}^{N \times (M\times L)}$ is obtained by copying $\boldsymbol{A}$ horizontally $L$ times (i.e.,  $\boldsymbol{B}_{ex}^T$). Then, the results of the parallel computing are calculated through the sum and reshape operations.

\begin{equation}
 \boldsymbol{Y}_{ex}  = 
 \begin{bmatrix}
   y_{11} & \cdots & y_{11} \\
   \vdots & \ddots & \vdots \\
   y_{M1} & \cdots & y_{M1} \\
   \hline
   y_{12} & \cdots & y_{12} \\
   \vdots & \ddots & \vdots \\
   y_{M2} & \cdots & y_{M2} \\
   \hline
   \vdots & \vdots & \vdots \\
   \vdots & \vdots & \vdots \\
   \hline
   y_{1L} & \cdots & y_{1L} \\
   \vdots & \ddots & \vdots \\
   y_{ML} & \cdots & y_{ML} \\
  \end{bmatrix}
  ,\boldsymbol{B}_{ex} = 
 \begin{bmatrix}
   {\Phi}_{11} & \cdots & {\Phi}_{1N} \\
   \vdots & \ddots & \vdots \\
   {\Phi}_{M1} & \cdots & {\Phi}_{MN}\\
   \hline
   {\Phi}_{11} & \cdots & {\Phi}_{1N} \\
   \vdots & \ddots & \vdots \\
   {\Phi}_{M1} & \cdots & {\Phi}_{MN}\\
   \hline
   \vdots & \vdots & \vdots \\
   \vdots & \vdots & \vdots \\
   \hline
   {\Phi}_{11} & \cdots & {\Phi}_{1N} \\
   \vdots & \ddots & \vdots \\
   {\Phi}_{M1} & \cdots & {\Phi}_{MN}\\
  \end{bmatrix}
  \notag
\end{equation}
\begin{equation}\label{6}
 \boldsymbol{Y}_{ex}\otimes\boldsymbol{B}_{ex}=
 \begin{bmatrix}
\boldsymbol{Y}_1\boldsymbol{B};
\boldsymbol{Y}_2\boldsymbol{B};
\cdots;
\cdots;
\boldsymbol{Y}_L\boldsymbol{B}
  \end{bmatrix}
\end{equation}

\begin{equation}\label{7}
\boldsymbol{X}_{ex}= \\
 \begin{bmatrix}
   x_{11} & \cdots & x_{11} &\cdots & x_{1L} & \cdots & x_{1L}\\
   \vdots & \cdots & \vdots  &  \cdots & \vdots & \cdots & \vdots\\
 x_{N1} & \cdots & x_{N1} & \cdots  & x_{NL} & \cdots & x_{NL}
  \end{bmatrix}
\end{equation}

In this way, the time complexity of each layer for DeepFPC-{$\ell_2$} is reduced from $O(n^3)$ to $O(n^2)$, and the extended matrix effectively improves the network construction speed.

\section{Experimental  Results}
\subsection{Sparse Signal Reconstruction}\label{train_l2}
The performance of the DeepFPC-{$\ell_2$} model is assessed in this section. The model is implemented using TensorFlow in Python and trained with the ADAM optimizer using an exponentially decaying step-size. A set of ${\lbrace}{\boldsymbol{y}^{l}}, {\boldsymbol{x}^{l}}{\rbrace}$ ($l = 1,2,...,L$) pairs ($L$ = 100) is used for training and a different set of 100 pairs is used for testing. We set the signal dimension to $N = 100$, and its sparsity level to $K = 10$. Moreover, we choose the support locations from the uniform distribution. Their values are drawn $i.i.d$ from the standard normal and the measurement matrix $\boldsymbol{\Phi}$ is randomly drawn from the Gaussian distribution $\mathcal{N}(0,1/M)$. The measurement vectors' dimension is $M = 300$ and are generated by computing (\ref{measurement}) with the signal and $\boldsymbol{\Phi}$. The normalized mean square error (NMSE), defined as NMSE(dB) $\triangleq$ $10\log_{10}\left({\left\|{{\boldsymbol{x}}^{\star} - \boldsymbol{x}}\right\|}_{2}^{2}/{\left\|{\boldsymbol{x}}\right\|}_{2}^{2}\right)$, is used as the metric, with ${\boldsymbol{x}}^{\star}$ being the reconstructed vector. 

\begin{table}[t]
\caption{Performance comparison between the FPC-{$\ell_2$} Algorithm and the proposed DeepFPC-{$\ell_2$} Network.}
\begin{center}
\setlength{\tabcolsep}{0.2mm}{
\begin{tabular}{ccc|ccc}
\hline
\textbf{\begin{tabular}[c]{@{}c@{}}Layers/\\ Iterations\end{tabular}} & \textbf{\begin{tabular}[c]{@{}c@{}}FPC-\textit{$l_2$} \\ $[dB]$\end{tabular}} & \textbf{\begin{tabular}[c]{@{}c@{}}DeepFPC-\textit{$l_2$}\\ $[dB]$\end{tabular}} & \textbf{\begin{tabular}[c]{@{}c@{}}Layers/\\ Iterations\end{tabular}} & \textbf{\begin{tabular}[c]{@{}c@{}}FPC-\textit{$l_2$}\\ $[dB]$\end{tabular}} & \textbf{\begin{tabular}[c]{@{}c@{}}DeepFPC-\textit{$l_2$}\\ $[dB]$\end{tabular}} \\ \hline
\textbf{1} & -4.53 & -12.27 & \textbf{11} & -6.34 & -14.64 \\
\textbf{2} & -4.80 & -12.32 & \textbf{12} & -6.48 & -15.07 \\
\textbf{3} & -5.03 & -12.67 & \textbf{13} & -6.62 & -15.16 \\
\textbf{4} & -5.23 & -12.71 & \textbf{14} & -6.76 & -15.34 \\
\textbf{5} & -5.42 & -13.03 & \textbf{15} & -6.89 & -15.51 \\
\textbf{6} & -5.59 & -13.87 & \textbf{16} & -7.02 & -15.84 \\
\textbf{7} & -5.76 & -14.00 & \textbf{17} & -7.16 & -15.90 \\
\textbf{8} & -5.91 & -14.41 & \textbf{18} & -7.29 & -15.99 \\
\textbf{9} & -6.06 & -14.53 & \textbf{19} & -7.42 & -16.00 \\
\textbf{10} & -6.20 & -14.64 & \textbf{20} & -7.55 & -16.16 \\
\hline
\end{tabular}}
\label{tab1}
\end{center}
\end{table}
Table \ref{tab1} shows the performance comparison between the FPC-{$\ell_2$} algorithm and the DeepFPC-{$\ell_2$} network in terms of the averaged NMSE for the first 20 iterations/layers. It is clear that the DeepFPC-{$\ell_2$} model significantly improves the corresponding algorithm. The FPC-{$\ell_2$} algorithm will converge to -14.39dB after 150 iterations. However, after 8 layers, DeepFPC-{$\ell_2$} has achieved this performance and it continues to reduce the NMSE as the number of layers increases. In this 20-layer network, the best result is better than the original algorithm with about 2dB margin. 

\subsection{Reconstruction in the Noisy Scenario}\label{5B}
In this section, we compare DeepFPC-{$\ell_2$} with DeepFPC~\cite{xiao2019deepfpc} in terms of their performance in the presence of noise in the measurements. The same settings as Section \ref{train_l2} are designed for training the DeepFPC~\cite{xiao2019deepfpc} model. Two kinds of noise, i.e., noise resulting from the measurement and the transmission process, are considered.

\subsubsection{Measurements contaminated by Gaussian noise}
\begin{figure}[t]
\centerline{\includegraphics[scale = 0.4]{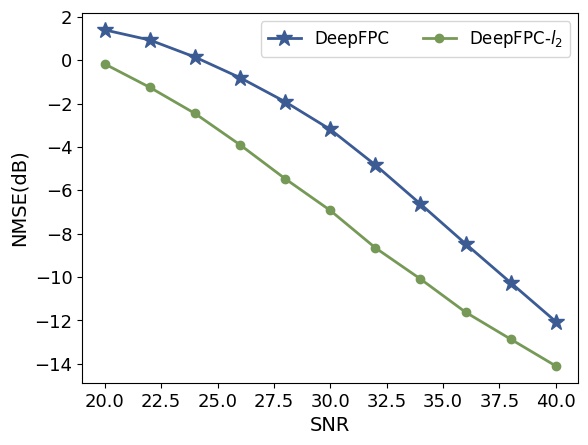}}
\caption{Comparison of the robustness to noise for DeepFPC~\cite{xiao2019deepfpc} and the proposed DeepFPC-{$\ell_2$} for SNR = 20dB to 40dB}
\label{fig.2}
\end{figure}

Unlike the training set, the measurements in the testing set are contaminated by Gaussian noise. The signal-to-noise ratio (SNR) in dB is defined as:
\begin{equation}
    SNR = 10\log_{10}\Big(\frac{P_S}{P_N}\Big),
\end{equation}
where $P_S$ and $P_N$ are the power of the signal and noise, respectively. The amplitude of noise $A_N$ can be calculated as:
\begin{equation}
    A_N = \sqrt{{P_S}/10^{(\frac{SNR}{10})}}
\end{equation}
The Gaussian noise follows the distribution $\mathcal{N}(0, A_N^2)$ and is added to the measurement matrix; different SNR levels are considered.  DeepFPC and DeepFPC-{$\ell_2$} are trained with a noiseless sampling set, and then they are tested with the noisy measurements. The results are shown in Fig.\ref{fig.2}. 
It is clear that the recovery performance of the proposed DeepFPC-{$\ell_2$} network is better than that of our previous DeepFPC model, which means that DeepFPC-{$\ell_2$} has better robustness to noise. 

Furthermore, we also tried to recover signals under low SNR conditions (less than 20dB). Although the NMSE of DeepFPC-{$\ell_2$} is still less than that of the DeepFPC, neither of them recover the signals with high accuracy. Improving the accuracy of DeepFPC-based model in the low SNR regime is still an open problem, which we plan to address in subsequent work.

\subsubsection{Measurements contaminated by transmission error}
\begin{figure}[t]
\centerline{\includegraphics[scale = 0.4]{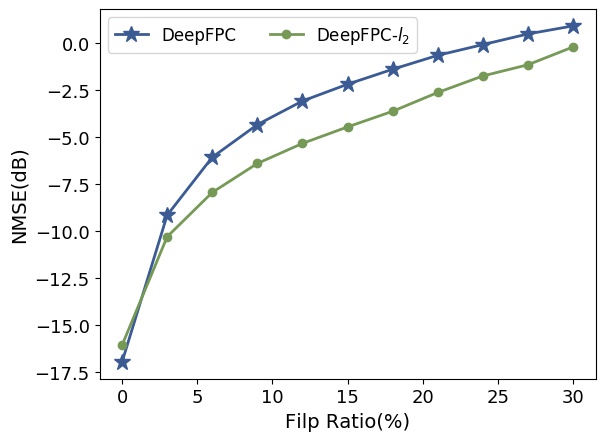}}
\caption{Comparison of the robustness to noise for DeepFPC and DeepFPC-{$\ell_2$} on ratio = 0\% to 30\%}
\label{fig.3}
\end{figure}

We now consider sign flips, which are caused by bit errors in the transmission process. In order to model sign flips, the measurement vector $\boldsymbol{y}$ is multiplied by a vector $\boldsymbol{\xi} \in {\lbrace}{-1,1}{\rbrace}^M$ to control flips. When an entry in the measurement vector is multiplies by -1, the sign of the entry flips. The locations of -1 are chosen randomly from 1 to 300 ($M = 300$) and the number of -1 is determined by the flip ratio (see Fig.\ref{fig.3}). 

In Fig.\ref{fig.3} we see that DeepFPC-{$\ell_2$} has a better signal reconstruction performance than the previous DeepFPC network when the flip ratio is equal to 3\% or higher; this shows that the DeepFPC-{$\ell_2$} has better robustness to noise than the DeepFPC. However, we can also notice that the recovery performance of the DeepFPC is better than the DeepFPC-{$\ell_2$} when the flip ratio is small enough.
This indicates that DeepFPC outperforms DeepFPC-$\ell_2$ in the noiseless scenario, and the reason could be that the associated one-sided {$\ell_1$}-norm algorithm has superior performance in the noiseless case compared to the one-sided {$\ell_2$}-norm\cite{jacques2013robust} algorithm. 

To sum up, the proposed DeepFPC-{$\ell_2$} network outperforms the traditional FPC-{$\ell_2$} algorithm for sparse signal recovery. Moreover, compared with the previous DeepFPC~\cite{xiao2019deepfpc} model, it has better robustness when the measurements are contaminated by noise, for a wide range of SNRs.

\section{Conclusion}
DeepFPC-{$\ell_2$}, a deep neural network designed by unfolding the FPC-{$\ell_2$} recovery algorithm for 1-bit CS, is proposed in this paper. Our result demonstrates that the DeepFPC-{$\ell_2$} not only improves the recovery performance of the original algorithm, but also greatly improves the convergence speed. Moreover, we explore the robust property of the DeepFPC-{$\ell_2$} network, and we find that the DeepFPC-{$\ell_2$} network has better noise immunity than the previous DeepFPC model. The results also indicate that the robust property of a deep unfolding network relies on the algorithm it stems from. When the measurements are polluted by noise, DeepFPC-{$\ell_2$} is a better option to obtain more accurate results. This approach can be applied to many 1-bit CS applications which can be affected by noise, such as DOA estimation, image processing and wireless communications.\vspace{-0.5em}

\section*{Acknowledgment}
This research received funding from the Flemish Government under the “Onderzoeksprogramma Artificiële Intelligentie (AI) Vlaanderen” programme and from the FWO (Projects G040016N and G0A4720N). The work was also supported by the Natural Science Foundation of China (grant no. 61901273).

\bibliographystyle{IEEEtran}
\bibliography{IEEEabrv,reference}
\end{document}